\documentclass[letterpaper, 10 pt, conference]{ieeeconf}  

\IEEEoverridecommandlockouts                              

\usepackage{times} 
\usepackage{graphicx}
\usepackage{amsmath} 
\usepackage{amssymb}  
\usepackage{algorithm}
\usepackage{algorithmic}

\usepackage{float}
\usepackage{subfig}

\title{\LARGE \bf
Evolving the Complete Muscle: Efficient Morphology-Control Co-design for Musculoskeletal Locomotion
}

\author{Lidong Sun$^{1}$, Wentao Zhao$^{2}$, Ye Wang$^{3}$, Huaping Liu$^{2}$, and Fuchun Sun$^{2}$
\thanks{Corresponding author: Fuchun Sun.}%
\thanks{$^{1}$Lidong Sun is with Naval University of Engineering, Wuhan 430033, China
        {\tt\small 13942645995@163.com}}%
\thanks{$^{2}$Wentao Zhao, Huaping Liu, Fuchun Sun are with the Department of Computer Science and Technology, Tsinghua University, Beijing 100084, China
        {\tt\small zhaowt25@mails.tsinghua.edu.cn, hpliu@tsinghua.edu.cn, fcsun@mail.tsinghua.edu.cn}}%
\thanks{$^{3}$Ye Wang is with the Department of Mechanical Engineering, Tsinghua University, Beijing 100084, China
        {\tt\small Alexye.wang22@gmail.com}}%
}

\begin{document}

\maketitle
\thispagestyle{empty}
\pagestyle{empty}

\begin{abstract}

Musculoskeletal robots offer intrinsic compliance and flexibility, providing a promising paradigm for versatile locomotion. However, existing research typically relies on models with fixed muscle physiological parameters. This static physical setting fails to accommodate the diverse dynamic demands of complex tasks, inherently limiting the robot's performance upper bound. In this work, we focus on the morphology and control co-design of musculoskeletal systems. Unlike previous studies that optimize single physiological attributes such as stiffness, we introduce a Complete Musculoskeletal Morphological Evolution Space that simultaneously evolves muscle strength, velocity, and stiffness. To overcome the exponential expansion of the exploration space caused by this comprehensive evolution, we propose Spectral Design Evolution (SDE), a high-efficiency co-optimization framework. By integrating a bilateral symmetry prior with Principal Component Analysis (PCA), SDE projects complex muscle parameters onto a low-dimensional spectral manifold, enabling efficient morphological exploration. Evaluated on the {{MyoSuite}} framework across four tasks (Walk, Stair, Hilly, and Rough terrains), our method demonstrates superior learning efficiency and locomotion stability compared to fixed-morphology and standard evolutionary baselines.

\end{abstract}

\section{INTRODUCTION}

Musculoskeletal robots, inspired by the intricate biomechanical structures of biological organisms, offer a promising paradigm to achieve biological-level agility and versatility in robotic locomotion. Unlike traditional robots driven by joint motors, these bio-inspired systems leverage muscle-tendon units (MTUs) which exhibit complex non-linear dynamics and intrinsic compliance \cite{zajac1989muscle, scott2004optimal}. In these systems, muscles serve as the primary drivers of movement, providing the essential flexibility and bio-fidelity required for interaction with the complex physical world \cite{caggiano2022myosuite}.

However, most existing research on musculoskeletal control typically employs models with fixed muscle physiological parameters. Such static physical settings inherently limit the upper bound of learning algorithms, as a fixed morphology restricts the robot's capacity to handle tasks with varying dynamic demands. In nature, biological organisms exhibit remarkable adaptability by evolving their physical structures alongside their nervous systems to better suit their environments \cite{pfeifer2006body, luck2020data}. Drawing inspiration from this biological phenomenon, it is imperative to move beyond pure control optimization and pursue the co-optimization of muscle morphological parameters and control policies.

\begin{figure}[t]
    \centering
    \includegraphics[width=\linewidth]{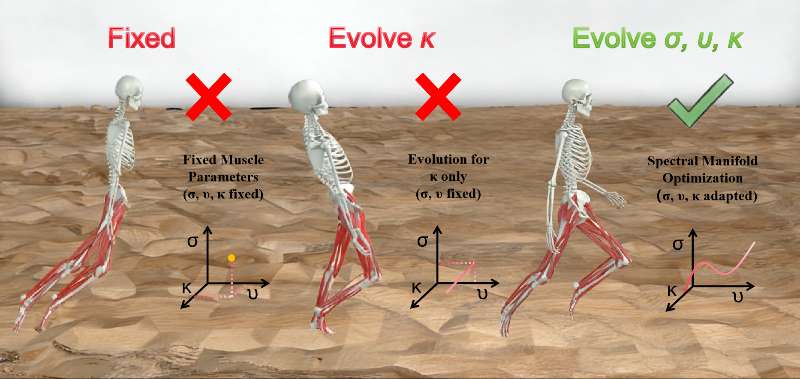}
    \caption{Conceptual overview of Spectral Design Evolution (SDE). (Left) Fixed-morphology agent fails on challenging terrains due to physical limits. (Right) SDE optimizes muscle triad parameters ($\sigma, \nu, \kappa$) on a low-dimensional spectral manifold, enabling the robot to adapt its physical capacity for superior locomotion stability.}
    \label{fig:motivation}
\end{figure}

A recent study~\cite{zhao2023bayesian} has explored muscle development by optimizing stiffness parameters to enhance task performance. While effective for specific scenarios like grasping, we observe that optimizing a single dimension such as stiffness is insufficient for highly dynamic tasks like locomotion, which primarily rely on active force generation. As illustrated in Fig.~\ref{fig:motivation}, if the system optimizes only passive stiffness without sufficient active strength, the robot may struggle to learn adaptive force-generation patterns required to traverse rough terrains or climb stairs. To address this, we propose a more comprehensive musculoskeletal morphological evolution space that simultaneously optimizes muscle \textbf{strength}, \textbf{velocity}, and \textbf{stiffness}.

While this multi-dimensional expansion provides the necessary physical foundation for complex behaviors, it inevitably introduces a severe ``curse of dimensionality''. This vast and functionally interdependent parameter space poses immense exploration challenges, causing standard reinforcement learning and co-optimization methods to suffer from low sample efficiency and a high susceptibility to suboptimal local minima. To mitigate this, we propose the \textit{Spectral Design Evolution} (SDE) paradigm (see Fig. \ref{fig:framework}). By coupling bilateral symmetry priors with principal component mapping, SDE reformulates the high-dimensional muscle design problem into a low-dimensional spectral manifold search, enabling the synchronous evolution of complex biological morphology and control policies.

The main contributions of this work are summarized as follows:
\begin{itemize}
\item \textbf{Complete Musculoskeletal Morphological Evolution Space:} We construct a comprehensive evolution space for musculoskeletal systems by incorporating a triad of critical physiological properties: strength, velocity, and stiffness, allowing the agent to transcend fixed physical limits and adapt to more diverse task scenarios.
\item \textbf{Spectral Design Evolution (SDE) Algorithm:} We propose a novel co-design algorithm that reformulates the high-dimensional muscle design problem into a low-dimensional spectral manifold search. By coupling bilateral symmetry priors with principal component mapping, SDE enables the synchronous evolution of complex biological morphology and control policies.
\item \textbf{Accelerated Convergence and Multidimensional Discovery:} We demonstrate through extensive experiments across four complex terrain scenarios that SDE leads to significantly faster learning convergence and superior stability compared to a fixed-morphology baseline and a direct-search co-design baseline. Notably, our findings reveal that the synergistic evolution of the three muscle parameters yields higher robustness than optimizing stiffness alone.
\end{itemize}

\section{Related Work}

\subsection{Musculoskeletal Control and Synergy}
Given the high-dimensional and overactuated nature of musculoskeletal robots, learning effective control policies has persistently been a significant challenge. Existing research mainly focuses on overcoming the control difficulties, proposing a variety of algorithms to simplify policy learning. One line of research seeks to augment exploration via well-designed noise mechanisms. For instance, DEP-RL~\cite{schumacher2022dep} generates action-correlated noise with Differential Extrinsic Plasticity, and Lattice~\cite{simpson2021data} applies temporally correlated noise to the policy's latent state to induce cross-actuator synergies. Another line of work leverages muscle synergy priors to facilitate the learning process. SAR~\cite{chiappa2024acquiring} transfers synergistic representations learned from simple tasks to complex tasks. DynSyn~\cite{he2024dynsyn} uses random perturbations to generate synergistic representations from the system's dynamical structure, achieving fast learning efficiency. Diff-Muscle~\cite{diffmuscle2026} exploits the theory of differential flatness to analytically map the redundant muscle activation space onto a low-dimensional joint space, enabling efficient policy learning for highly dynamic robotic table tennis tasks.

Although these methods significantly alleviate the learning burden for musculoskeletal systems, they universally rely on fixed-parameter models. By treating the morphology as a static constraint, existing research overlooks the potential to expand the robot's mechanical capabilities through morphological adaptation~\cite{seth2018opensim, peng2018deepmimic}. To break this physical bottleneck, our work shifts the focus toward the synergistic co-optimization of muscle physiological parameters alongside the control policy.

\subsection{Morphology-Control Co-design}
The co-design of morphology and control has gathered significant attention in recent years, as it simultaneously evolves morphological parameters and learns control policies to adapt to task-specific demands~\cite{sims1994evolving, liu2025embodied}. Transform2Act~\cite{yuan2022t2a} introduced a framework to modify skeletal attributes within the reinforcement learning loop. Subsequently, SARD~\cite{dong2023sard} proposed a symmetry-aware robot design method that utilizes structured subgroups to search for optimal symmetry, effectively reducing the complexity of the design space. Building on this, CompetEvo~\cite{huang2024competevo} incorporated morphological evolution into competitive scenarios by co-evolving agent designs and tactics in adversarial games. Recently, BodyGen~\cite{lu2025bodygen} advanced efficient embodiment co-design through topology-aware self-attention and temporal credit assignment mechanisms. Meanwhile, Bayesian~\cite{zhao2023bayesian} proposed Bayesian Morphology Optimization, which successfully evolves muscle parameters to enhance grasping performance. However, Bayesian Morphology Optimization simplifies the design space by evolving only a single physiological parameter, specifically stiffness, under the assumption that passive compliance is the dominant factor in task success. In contrast, dynamic locomotion across varied terrains requires a more holistic evolution of the muscle-tendon triad, comprising strength, velocity, and stiffness. Concurrently evolving these parameters in a high-dimensional space leads to severe efficiency bottlenecks, which our proposed Spectral Design Evolution framework addresses by reformulating the design problem into a low-dimensional spectral manifold search.

\section{Preliminary}

\subsection{Problem Formulation}
We formulate the musculoskeletal morphology-control co-optimization problem as an augmented Markov Decision Process (MDP) defined by a 7-element tuple $\mathcal{M} = \langle \mathcal{S}, \mathcal{A}, \mathcal{D}, \Phi, \mathcal{T}, \mathcal{R}, \gamma \rangle$, where $\mathcal{S} \subseteq \mathbb{R}^n$ represents the continuous state space, $\mathcal{A} \subseteq \mathbb{R}^{d+e}$ represent the actions for morphology design and muscle control conditioned on stage, and $\mathcal{D} \subseteq \mathbb{R}^{d}$ is the muscle parameter design space. $\Phi$ is a discrete stage flag indicates whether the agent is currently in the process of evolving its morphology or executing physical movements. At design stage, the agent samples a morphological design action $a_t^d$ from design policy $\pi_\theta$ based on the current state $s_t$ to change the morphology:
\begin{equation}
    a_t^d \sim \pi_\theta(\cdot | s_t).
\end{equation}
At control stage, the agent samples an action $a_t^e$ from control policy $\pi_\phi$ which indicates the neural excitation:
\begin{equation}
    a_t^e \sim \pi_\phi(\cdot | s_t).
\end{equation}
$\mathcal{T}(s_{t+1}|s_t,a_t^d,a_t^e)$ is the transition function and $\mathcal{R}: \mathcal{S} \times \mathcal{A} \rightarrow \mathbb{R}$ is the reward function.

The goal is to jointly learn the optimal design policy $\pi_\theta$ and control policy $\pi_\phi$ which maximizes the expected total discounted return:
\begin{equation}
    J(\pi_\theta,\pi_\phi) = \mathbb{E}_{\tau \sim \pi_\theta, \pi_\phi} \left[ \sum_{t=0}^{\infty} \gamma^t r(s_t, a_t) \right],
\end{equation}
where $\gamma \in [0,1)$ is discount factor, and $T$ is the episode horizon. We optimize the policy with Proximal Policy Optimization (PPO)~\cite{schulman2017proximalpolicyoptimizationalgorithms}.

\subsection{Musculoskeletal Modeling and Physiological Parameterization}
In this paper, we use the musculoskeletal models implemented in the MuJoCo physics simulator, where the muscle force $f_m$ is formulated as follows:
\begin{equation}
    f_m = -(F_L(L) \cdot F_V(V) \cdot a + F_P(L)) \cdot F_0
\end{equation}
where $F_L(L)$ denotes the active force-length function, $F_V(V)$ denotes the force-velocity function, and $F_P(L)$ denotes the passive force that depends solely on the muscle length. $a$ is the muscle activation value and $F_0$ is the peak force at rest length. 

To enable morphological co-optimization, we parameterize the Hill-type muscle model across three critical physiological dimensions: strength ($\sigma$), velocity ($\nu$), and stiffness ($\kappa$). These parameters reshape the dimensionless characteristic curves that govern the instantaneous muscle force $f_m$.

\begin{itemize}
    \item \textbf{Strength ($\sigma$):} 
    The peak isometric force $F_0$ is physiologically determined by the physiological cross-sectional area (PCSA). We introduce $\sigma$ as a linear scaling factor to the reference peak force $F_{0}^{ref}$:
    \begin{equation}
        F_0(\sigma) = \sigma \cdot F_{0}^{ref}
    \end{equation}
    where $\sigma$ acts as a gain on the muscle's output magnitude. Increasing $\sigma$ enhances the maximum load-bearing capacity without altering the temporal or restorative dynamics of the actuator.

    \item \textbf{Velocity ($\nu$):} 
    The force-velocity relationship $F_V$ characterizes the decline in force during muscle shortening. Let $v$ be the absolute contraction velocity and $v_{max}^{ref}$ be the default intrinsic maximum velocity. We parameterize the normalized velocity $\tilde{v}$ using the scaling factor $\nu$:
    \begin{equation}
        \tilde{v} = \frac{v}{\nu \cdot v_{max}^{ref}}
    \end{equation}
    Substituting this into the active force component, the force-velocity multiplier $F_V(\tilde{v})$ modulates the force output. 
    Physiologically, $\nu$ represents the fiber-type composition (e.g., fast-twitch vs. slow-twitch fibers). A higher $\nu$ increases the muscle's functional bandwidth, allowing it to maintain substantial force output at higher shortening speeds, which is essential for explosive, high-power movements.

    \item \textbf{Stiffness ($\kappa$):} 
    The passive force $F_P$ represents the restorative tension provided by parallel elastic elements (e.g., connective tissues and titin). Following the nonlinear elastic formulation \cite{millard2013flexing}, $F_P$ is defined by its exponential curvature $\kappa$:
    \begin{equation}
        F_P(L, \kappa) = \max \left( 0, \frac{e^{\kappa \cdot (L - 1)} - 1}{e^{\kappa \cdot (L_{max} - 1)} - 1} \right) 
    \end{equation}
    where $L$ is the normalized muscle length. Here, $\kappa$ serves as the passive stiffness coefficient. A larger $\kappa$ dictates a sharper rise in resistive force as the muscle is stretched beyond its rest length ($L > 1$). This parameterization allows the musculoskeletal system to exploit passive stability, by optimizing $\kappa$, the robot can achieve energy-efficient posture maintenance and joint limit protection through biomechanical impedance rather than active neural excitation.
\end{itemize}



\section{Methodology}

\begin{figure*}[t] 
    \vspace{10pt}
    \centering
    \includegraphics[width=0.9\textwidth]{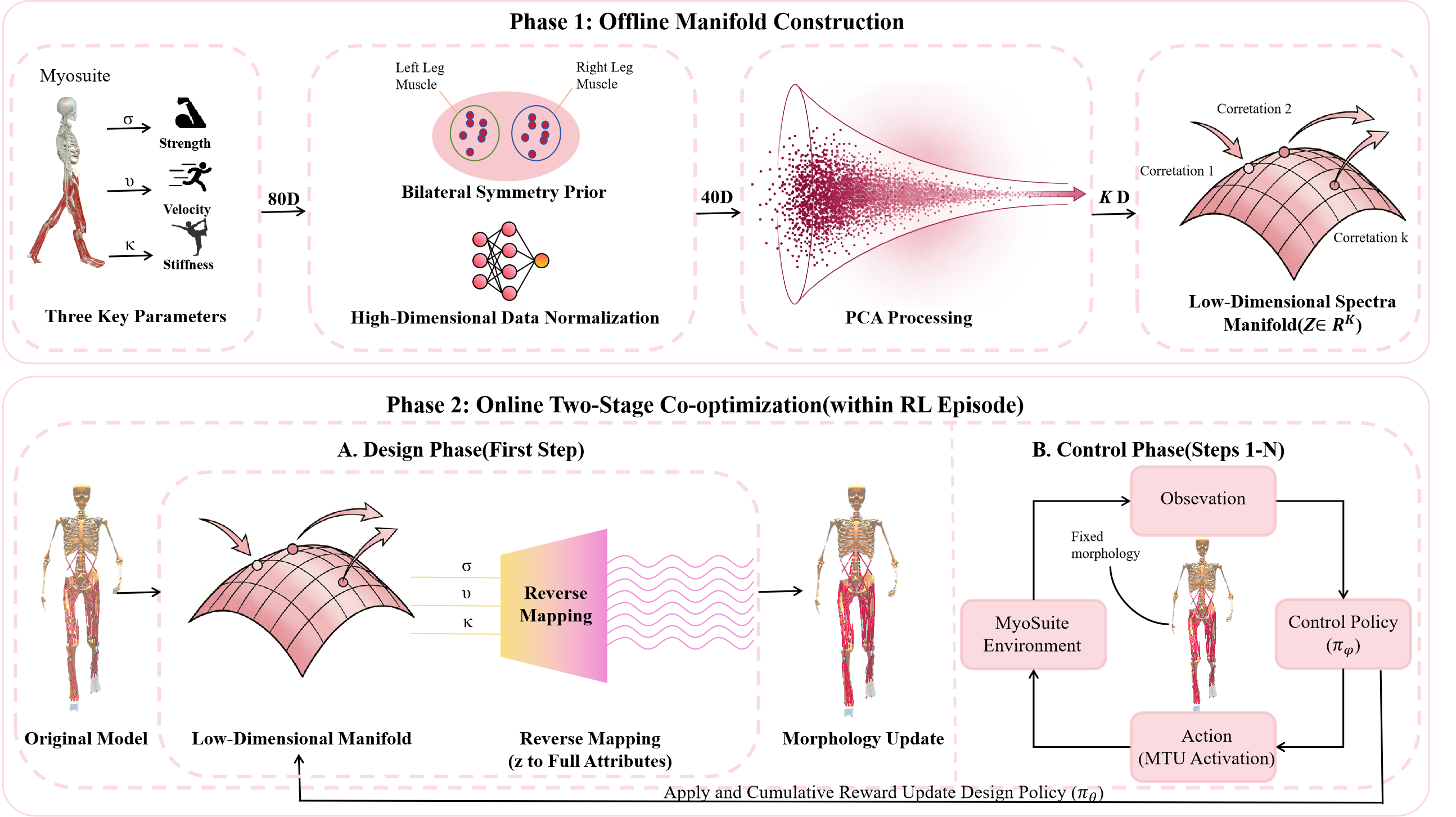}
    \caption{The SDE framework architecture. The process consists of an offline dynamics-aware manifold construction phase and an online two-stage co-optimization phase where the design policy $\pi_\theta$ and control policy $\pi_\phi$ are optimized via shared reward signals.}
    \label{fig:framework}
\end{figure*}

In this section, we present the \textit{Spectral Design Evolution} (SDE) framework. The core idea is to project the high-dimensional, highly redundant muscle parameter space into a low-dimensional spectral manifold, followed by a two-stage co-optimization process for morphology and control. The overall architecture of the SDE framework is depicted in Fig.~\ref{fig:framework}. It comprises an offline phase for spectral manifold construction and an online phase for two-stage co-optimization, where the design policy $\pi_\theta$ and control policy $\pi_\phi$ are trained synchronously to adapt to diverse terrains. The overall procedure is summarized in Algorithm~\ref{alg:sde}.

\begin{algorithm}[h]

\caption{Spectral Design Evolution (SDE)}
\label{alg:sde}
\begin{algorithmic}[1]
\STATE \textbf{Offline Phase: Dynamics-Aware Manifold Construction}
\STATE \quad Execute random exploration to collect muscle length history $\mathbf{H}$.
\STATE \quad Apply Symmetry Grouping and record group-averaged lengths.
\STATE \quad Perform Standardization and PCA to compute kinematic basis $\mathbf{V}_k \in \mathbb{R}^{M \times k}$.
\STATE \quad \textbf{Pre-compute} expanded basis $\tilde{\mathbf{V}}_k \in \mathbb{R}^{3M \times 3k}$ (block-diagonal of $\mathbf{V}_k$).
\STATE \textbf{Online Phase: Two-Stage Co-optimization}
\STATE \quad Initialize design policy $\pi_\theta$ and control policy $\pi_\phi$.
\FOR{episode $e = 1, \dots, E$}
    \STATE \textit{\textbf{Stage 1: Morphological Design ($t=0$)}}
    \STATE \quad Sample latent code $\mathbf{z} \in \mathbb{R}^{3k} \sim \pi_\theta(\mathbf{z} | s_0)$.
    \STATE \quad \textbf{Compute} morphology $\Theta = \bar{\Theta} + \tilde{\mathbf{V}}_k \mathbf{z}$. 
    \STATE \quad \textbf{Apply} $\Theta$ to MuJoCo XML parameters ($\sigma, \nu, \kappa$).
    \STATE \textit{\textbf{Stage 2: Locomotion Control ($t=1 \dots T$)}}
    \FOR{$t = 1, \dots, T$}
        \STATE Sample action $\mathbf{a}_t \sim \pi_\phi(\mathbf{a}_t | s_t, \Theta)$ and execute in MyoSuite.
    \ENDFOR
    \STATE Update $\theta$ and $\phi$ using PPO based on cumulative reward $R$.
\ENDFOR
\end{algorithmic}
\end{algorithm}

\subsection{Complete Evolution Space Construction}
\label{subsec:complete_space}
To enable full-body morphological adaptation, we first define a \textbf{Complete Evolution Space} $\Omega$ that encompasses all critical physiological traits of the musculoskeletal system. For a robot model with $M$ symmetry-grouped muscles, the morphology is represented by a high-dimensional vector $\Theta \in \mathbb{R}^{3M}$. Each muscle group $i$ is characterized by its own triad $\theta_i = (\sigma_i, \nu_i, \kappa_i)$, which directly modifies the underlying MuJoCo simulation parameters:

\begin{itemize}
    \item \textbf{Implementation Details:} We target the \texttt{muscle} type actuators in the MuJoCo XML. The strength $\sigma$ scales the \texttt{gainprm[2]} (Peak Force), the velocity $\nu$ scales the \texttt{gainprm[6]} (Max Velocity), and the stiffness $\kappa$ modifies the \texttt{biasprm[2]} (Passive Force Shape).
    \item \textbf{Evolutionary Bounds:} To ensure biomechanical validity and prevent simulation instability, we define a bounding box for each parameter. We set $\sigma \in [0.5, 1.5]$ and $\nu \in [0.5, 1.5]$ as multipliers of the default values. For the stiffness shape parameter $\kappa$, we allow a range of $[0.5, 2.0]$ to account for the exponential sensitivity of passive tissues.
\end{itemize}

\subsection{Dynamics-Aware Spectral Design Space Construction}
\label{subsec:spectral_construction}
As the number of muscles $M$ increases (e.g., $M=40$ for a bipedal robot), the dimension of $\Omega$ reaches $120+$, creating a severe ``curse of dimensionality'' for standard RL. This motivates our spectral manifold approach.
Distinct from discrete grouping approaches, we utilize Principal Component Analysis (PCA) to construct a \textbf{continuous spectral manifold} that captures the underlying functional coupling of the musculoskeletal system. We posit that muscles exhibiting strong kinematic correlations should also follow coherent evolutionary constraints. 
This construction process corresponds to the \textbf{Offline Phase} (Lines 1--5) in Algorithm~\ref{alg:sde}.

\subsubsection{Symmetry-Informed Grouping}
To maintain biomechanical plausibility and reduce exploration dimensionality, we incorporate a \textbf{bilateral symmetry prior}. For a bipedal robot, we enforce $\theta_{\text{left}, i} = \theta_{\text{right}, i}$. This ensures that the evolved morphology remains balanced, preventing asymmetric gait patterns and ensuring stable forward locomotion.
\subsubsection{Dynamic Response Data Collection \& Decomposition}
We record symmetry-group averaged muscle lengths $\bar{L}_{t, i}$ under random excitation $\mathbf{a}_t \sim \mathcal{U}(0, 1)$. The resulting standardized history matrix $\tilde{\mathbf{H}} \in \mathbb{R}^{T \times M}$ captures the musculoskeletal manifold. By performing spectral decomposition on the covariance matrix $\mathbf{C} = \frac{1}{T-1} \tilde{\mathbf{H}}^\top \tilde{\mathbf{H}}$, we obtain the eigenvectors $\mathbf{v}_i$.

\subsubsection{Spectral Manifold Mapping}
Muscle properties in biological systems are coupled through functional synergies~\cite{d2003combinations}. Utilizing PCA~\cite{santello1998postural}, we construct the kinematic projection matrix $\mathbf{V}_k = [\mathbf{v}_1, \dots, \mathbf{v}_k] \in \mathbb{R}^{M \times k}$ from the top $k$ eigenvectors. This basis captures the principal directions of functional muscle coordination, allowing the design policy to search within a compact, low-dimensional latent space $\mathbf{z} \in \mathbb{R}^{3k}$ ($k \ll M$) rather than individual parameters, significantly accelerating convergence.

\subsection{Two-Stage Co-optimization Strategy}
\label{subsec:co_optimization}
We decouple the co-optimization into a hierarchical sequence within each episode. As summarized in Algorithm~\ref{alg:sde}, this strategy implements the \textbf{Online Phase} of the SDE framework (Lines 6--16), where each episode is divided into two functional stages: a one-shot \emph{Morphological Design} stage (Lines 9--12) and a multi-step \emph{Locomotion Control} stage (Lines 13--16). Within this online phase, we adopt a design-before-control paradigm similar to Transform2Act, where the body configuration is first determined and then exploited by the control policy throughout the remaining rollout. Different from Transform2Act, which focuses on modifying skeletal structures and body layouts, our setting keeps the kinematic topology fixed and instead evolves continuous muscle physiological parameters on the spectral manifold. This formulation preserves the benefit of episodic design--control coupling while remaining naturally suited to musculoskeletal systems.


\subsubsection{Phase 1: Morphological Design}
At $t=0$, the design policy $\pi_\theta$ observes the initial state $s_0$ and samples a latent code $\mathbf{z} \in \mathbb{R}^{3k} \sim \pi_\theta(\mathbf{z} | s_0)$. We employ an \textbf{expanded block-diagonal basis} $\tilde{\mathbf{V}}_k \in \mathbb{R}^{3M \times 3k}$ to map this latent code to the full physiological space:
\begin{equation}
    \Theta(\mathbf{z}) = \bar{\Theta} + \begin{bmatrix}
        \mathbf{V}_k & \mathbf{0} & \mathbf{0} \\
        \mathbf{0} & \mathbf{V}_k & \mathbf{0} \\
        \mathbf{0} & \mathbf{0} & \mathbf{V}_k
    \end{bmatrix}
    \begin{bmatrix}
        \mathbf{z}_\sigma \\
        \mathbf{z}_\nu \\
        \mathbf{z}_\kappa
    \end{bmatrix},
    \label{eq:morph_mapping}
\end{equation}
where $\mathbf{z}_\sigma, \mathbf{z}_\nu, \mathbf{z}_\kappa \in \mathbb{R}^k$. This construction ensures that strength, velocity, and stiffness are evolved through separate latent coordinates, while all three remain constrained by the same biomechanical synergy basis $\mathbf{V}_k$.

The resulting morphology vector $\Theta=(\boldsymbol{\sigma},\boldsymbol{\nu},\boldsymbol{\kappa})$ is then applied to the MuJoCo muscle parameters before any physical interaction begins. By restricting the design action to a one-shot decision at the episode boundary, SDE avoids intra-episode non-stationarity in the body dynamics. As a result, the downstream controller interacts with a fixed morphology throughout the rollout, which stabilizes policy optimization and makes the long-term effect of a sampled design easier to evaluate. Moreover, because the design policy searches in a low-dimensional latent space rather than directly perturbing all $3M$ physiological parameters, it can focus on coordinated body-level adaptations instead of noisy per-muscle modifications.

\subsubsection{Phase 2: Locomotion Control}
Once $\Theta$ is fixed, the remainder of the episode becomes a standard locomotion problem conditioned on the instantiated morphology. The control policy $\pi_\phi$ generates muscle activations $\mathbf{a}_t$ at each step $t$:
\begin{equation}
    \mathbf{a}_t \sim \pi_\phi(\mathbf{a}_t | s_t, \Theta),
    \qquad t=1,\dots,T.
\end{equation}
Conditioning the controller on $\Theta$ is important for two reasons. First, it allows a single policy to adapt its actuation strategy to different evolved bodies, rather than requiring an independent controller for every candidate morphology. Second, it enables experience sharing across nearby designs on the spectral manifold, thereby improving sample efficiency during joint training. In practice, the controller learns not only \emph{how to move}, but also \emph{how to exploit} different strength--velocity--stiffness configurations for terrain-dependent locomotion.

After the rollout terminates, the cumulative reward is used to jointly update $\pi_\theta$ and $\pi_\phi$ with PPO. The design action at $t=0$ receives an episode-level credit signal reflecting the overall quality of the instantiated morphology, whereas the control actions receive dense temporal credit along the trajectory. This asymmetric yet shared optimization structure naturally matches the roles of the two policies: $\pi_\theta$ performs global, low-frequency body adaptation, while $\pi_\phi$ performs local, high-frequency motor coordination. By reducing the morphological search space from $3M$ to $3k$ (e.g., from 240 to 15), SDE enables the efficient discovery of task-optimal bodies that substantially enhance locomotion stability and final task return.

\section{Experiments}
In this section, we evaluate the performance of the proposed Spectral Design Evolution (SDE) framework through a series of locomotion tasks. Our evaluation focuses on answering the following research questions:
\begin{itemize}
    \item \textbf{RQ1 (Efficiency):} Does SDE facilitate faster convergence and higher task performance compared to fixed-morphology and other co-optimization baselines?
    \item \textbf{RQ2 (Synergy):} Is the co-evolution of the full physiological triad $(\sigma, \nu, \kappa)$ superior to optimizing individual parameters?
    \item \textbf{RQ3 (Symmetry):} How does the bilateral symmetry prior affect the physical plausibility and stability of evolved locomotion?
    \item \textbf{RQ4 (Scalability):} How does the latent dimension $k$ govern the trade-off between design expressivity and exploration efficiency?
\end{itemize}

\subsection{Experimental Setup}

\begin{figure}[t]
    \vspace{10pt}
    \centering
    \includegraphics[width=1.0\linewidth]{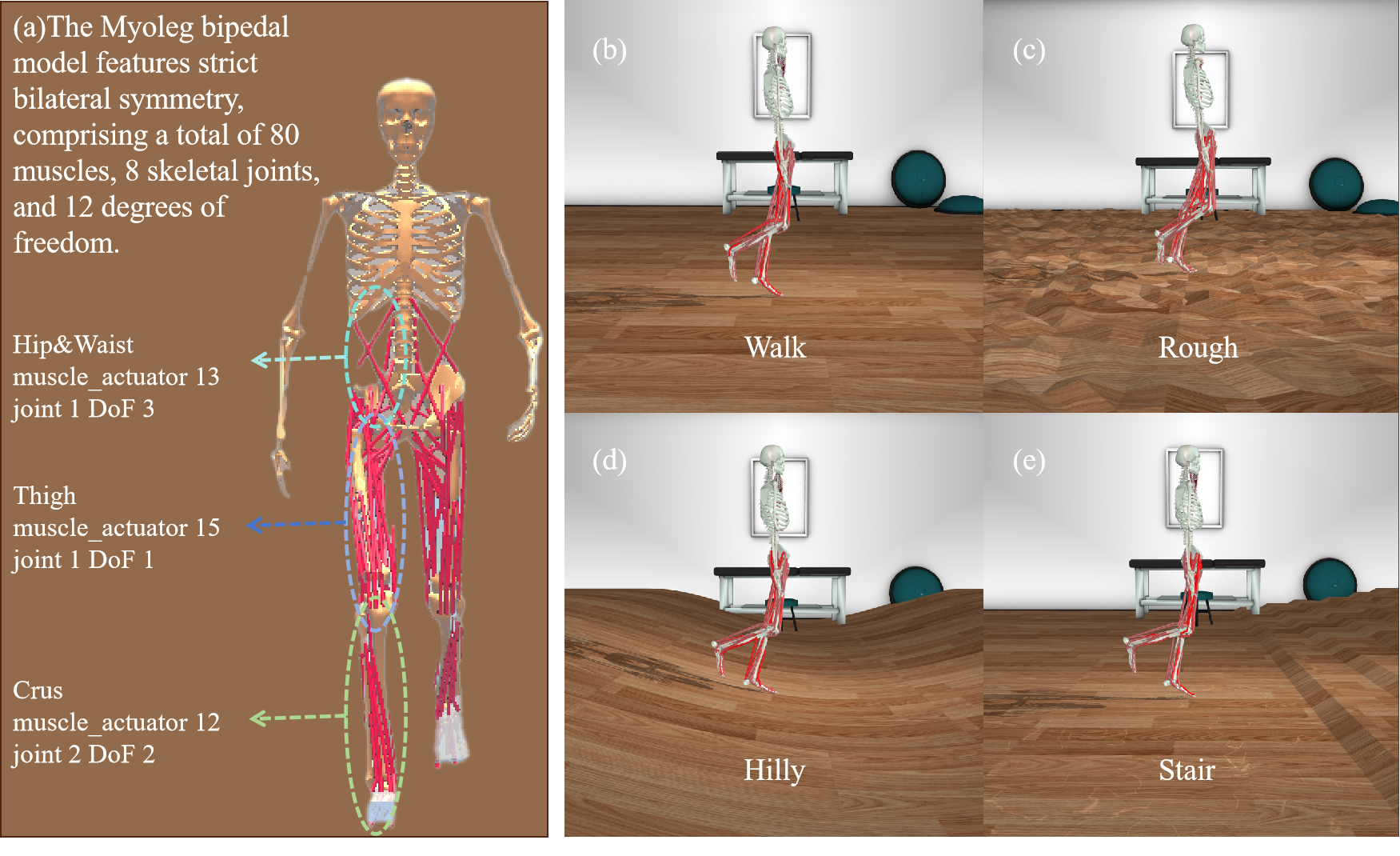}
    \caption{Experimental setup and task environments in MyoSuite: (a) The \textit{MyoLeg} model used for spectral analysis and co-optimization; (b) \textbf{Walk} terrain; (c) \textbf{Rough} terrain with stochastic debris; (d) \textbf{Hilly} terrain with undulating slopes; (e) \textbf{Stair} terrain with vertical obstacles.}
    \label{fig:setup}
\end{figure}

\subsubsection{Task Environments}
As illustrated in Fig.~\ref{fig:setup}, we utilize the \textit{MyoLeg} model (Fig.~\ref{fig:setup}(a)) within the MyoSuite~\cite{vittori2022myosuite}. The model features 80 muscles and is tasked with navigating four distinct landscapes that challenge different aspects of musculoskeletal dynamics:
\begin{itemize}
    \item \textbf{Walk (Fig.~\ref{fig:setup}(b)):} A flat terrain used to evaluate baseline cyclic gait stability and energy efficiency.
    \item \textbf{Rough (Fig.~\ref{fig:setup}(c)):} A terrain covered with stochastic debris, necessitating passive impact absorption and intrinsic compliance ($\kappa$).
    \item \textbf{Hilly (Fig.~\ref{fig:setup}(d)):} Undulating slopes that test the agent's ability to adapt to varying gravitational loads.
    \item \textbf{Stair (Fig.~\ref{fig:setup}(e)):} A series of vertical steps requiring high explosive power ($\sigma$) and precise foot placement.
\end{itemize}

\subsubsection{Baselines and Training}
We compare SDE with two baselines: 
\begin{itemize}
    \item \textbf{Fixed Morphology}, where only the control policy $\pi_\phi$ is trained on a default musculoskeletal model. 
    \item \textbf{Transform2Act (T2A)}, which performs direct $240$D morphological search without spectral mapping. 
\end{itemize}
All policies are optimized using PPO with a shared reward signal $R$ incorporating forward progress, balance maintenance, and energy effort penalties.

\subsection{Comparative Performance Analysis}
\begin{figure}[t]
    \centering
    \subfloat[Walk.]{
        \includegraphics[width=0.47\linewidth]{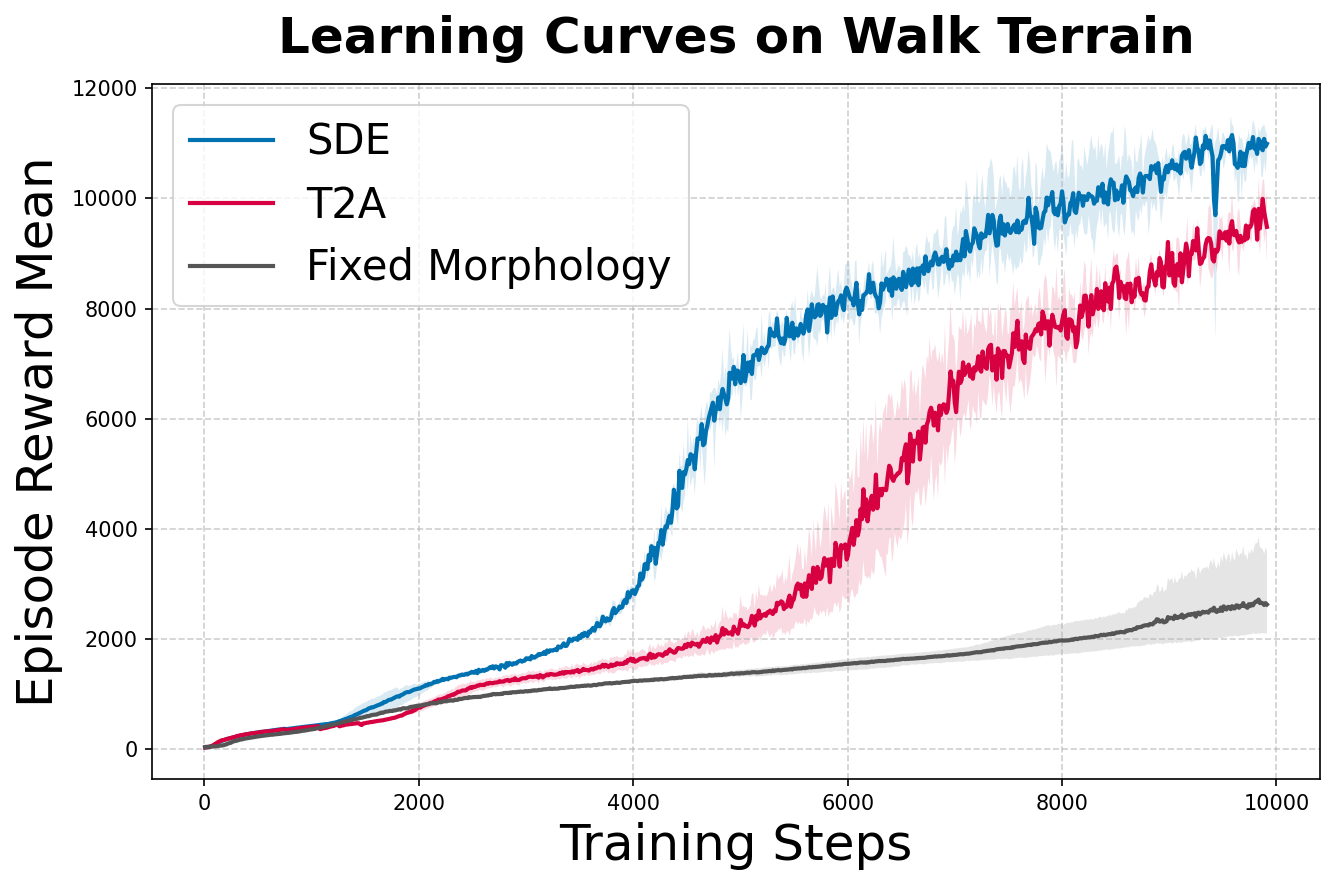}
        \label{fig:plot_walk}
    }
    \hfill
    \subfloat[Stair.]{
        \includegraphics[width=0.47\linewidth]{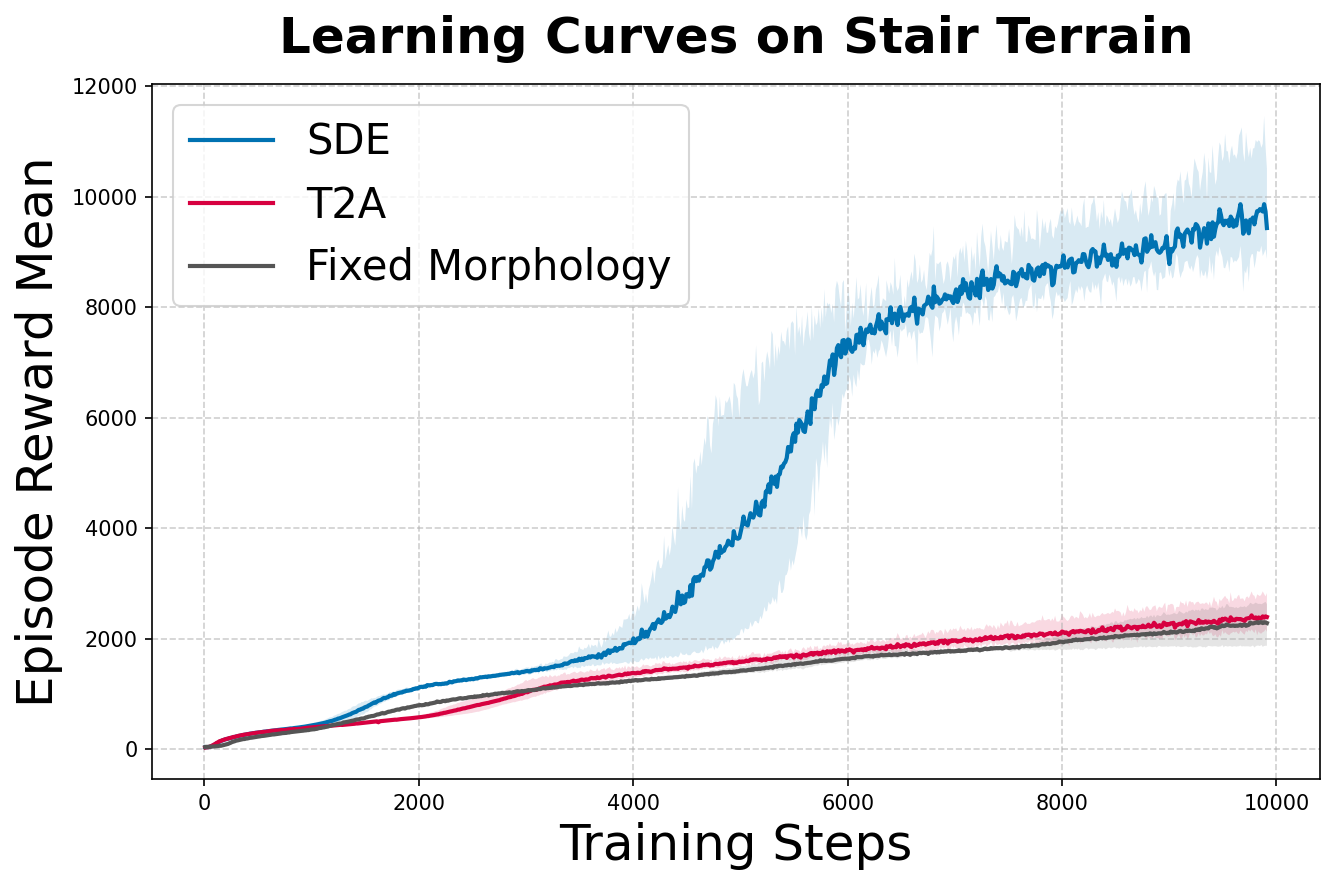}
        \label{fig:plot_stair}
    }
    \\
    \subfloat[Hilly.]{
        \includegraphics[width=0.47\linewidth]{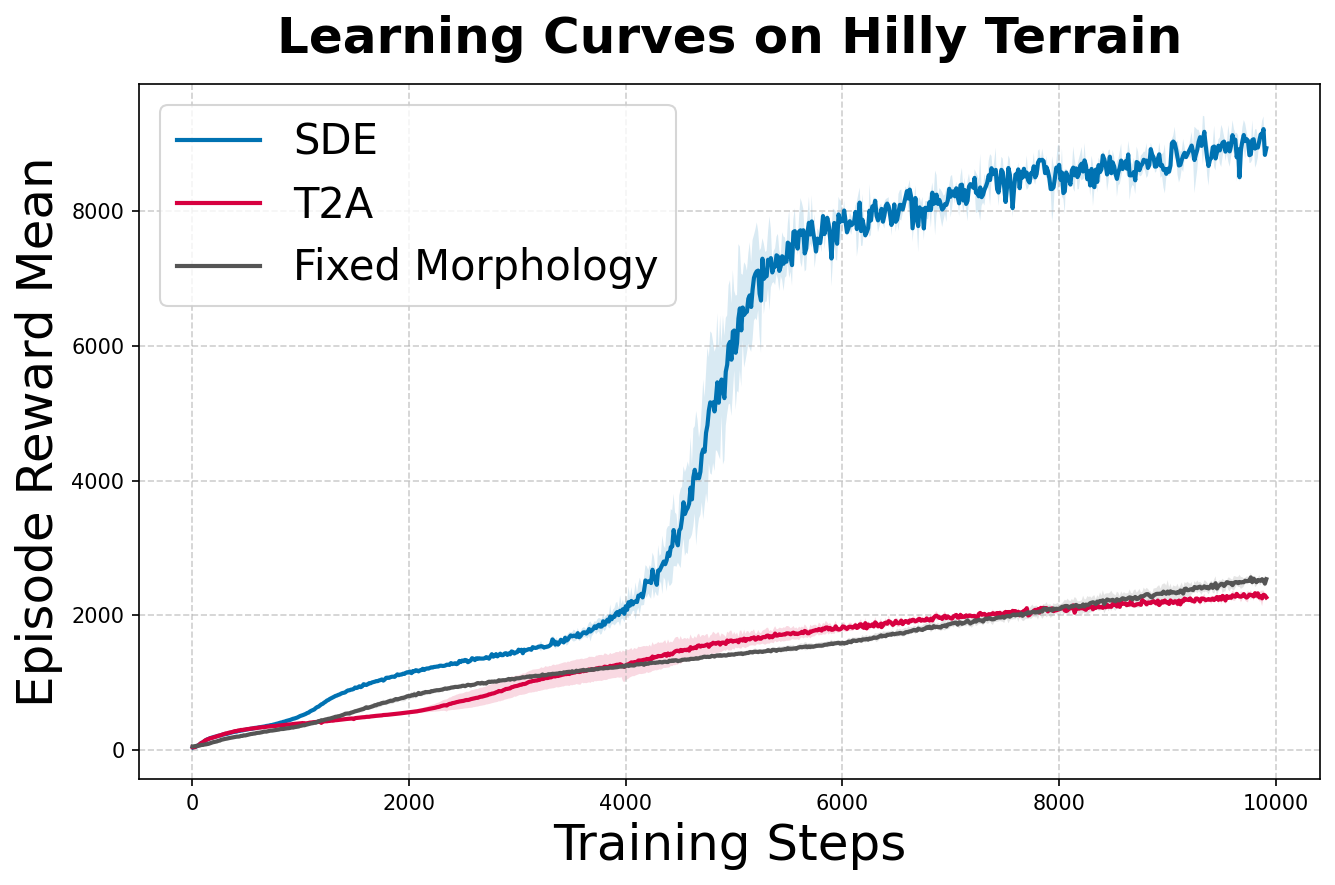}
        \label{fig:plot_hilly}
    }
    \hfill
    \subfloat[Rough.]{
        \includegraphics[width=0.47\linewidth]{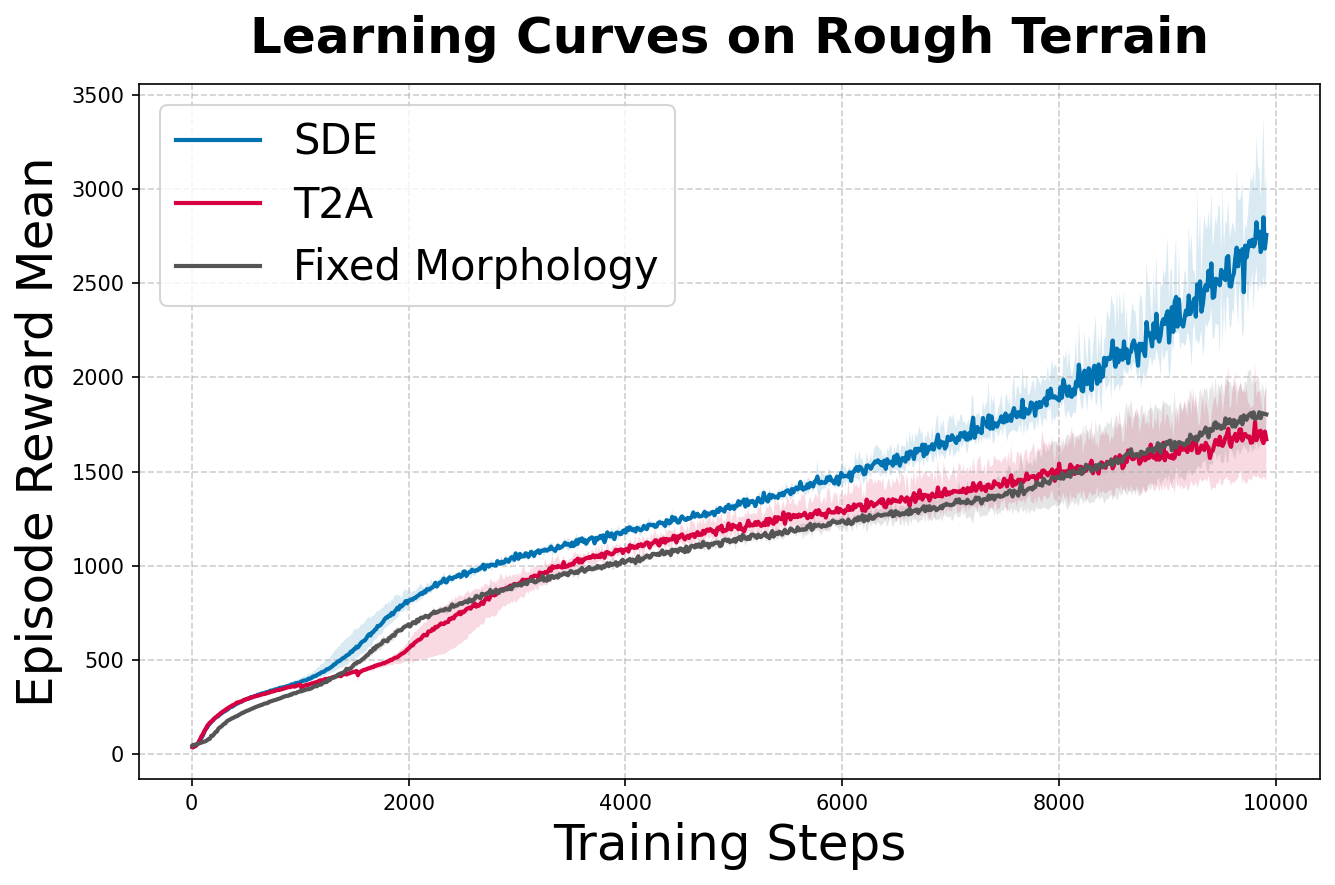}
        \label{fig:plot_rough}
    }
    \caption{Learning curves across four terrains. SDE demonstrates superior adaptability, particularly in high-impedance tasks like \textit{Stair}, \textit{Hilly} and \textit{Rough}.}
    \label{fig:learning_curves}
\end{figure}

As illustrated in Fig.~\ref{fig:learning_curves}, we evaluate the learning efficiency and final performance across four terrains. The results demonstrate that the SDE framework significantly outperforms both the \textit{Fixed} and \textit{T2A} baselines in terms of convergence speed and cumulative rewards.

Specifically, we observe a clear performance distinction based on task complexity:
\begin{itemize}
    \item \textbf{On Regular Terrain (Walk):} \textit{T2A} demonstrates competitive performance, eventually surpassing the \textit{Fixed} baseline. This suggests that in simple environments, direct co-optimization can eventually discover beneficial morphological adjustments.
    \item \textbf{On Complex Terrains (Stair, Hilly, Rough):} The performance of \textit{T2A} drops sharply, often remaining on par with the \textit{Fixed} baseline throughout the training duration. This indicates that \textit{T2A} is still struggling to explore the optimal morphological configuration within its unconstrained $240+$ dimensional design space.
\end{itemize}

In contrast, \textbf{SDE exhibits faster convergence across all scenarios.}
By searching within a low-dimensional spectral manifold, SDE substantially alleviates the exploration burden caused by the high-dimensional design space faced by \textit{T2A}.
This enables the agent to identify task-relevant biomechanical synergies more efficiently, transforming high-dimensional morphological search into a more tractable latent-space co-optimization problem.

\subsection{Ablation Studies}

\subsubsection{Physiological Triad Co-optimization}
To address \textbf{RQ2}, we evaluate the necessity of co-optimizing the full physiological triad $(\sigma, \nu, \kappa)$ in \textit{Rough} task by comparing it against variants where only a single parameter type is evolved: \textit{SDE-$\sigma$} (Strength only), \textit{SDE-$\nu$} (Velocity only), and \textit{SDE-$\kappa$} (Stiffness only).

\begin{figure}[t]
    \vspace{10pt}
    \centering
    \includegraphics[width=0.95\linewidth]{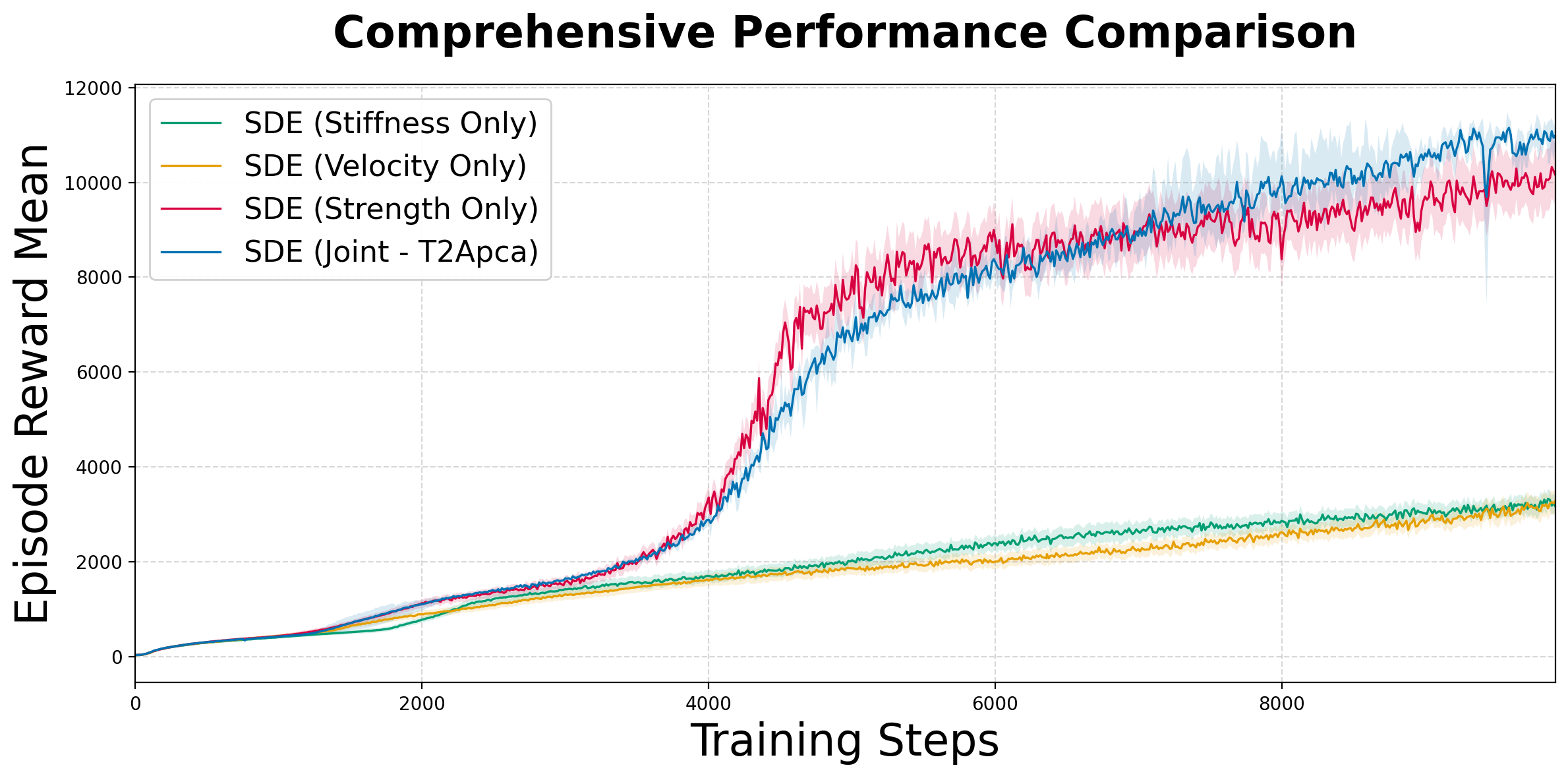}
    \caption{Learning curves comparing full triad co-optimization against single-parameter evolution. The synergy between parameters contributes to adaptation across diverse terrains.}
    \label{fig:triad_learning}
\end{figure}

As shown in Fig.~\ref{fig:triad_learning}, although single-parameter evolution provides marginal improvements over the \textit{Fixed} baseline, it fails to match the performance of the full SDE framework. Notably, while the \textit{SDE-$\sigma$} (Strength only) variant achieves a reward magnitude comparable to the full SDE framework, a qualitative analysis of the synthesized gait reveals a critical discrepancy. 

\begin{figure}[t]
    \centering
    \includegraphics[width=1.0\linewidth]{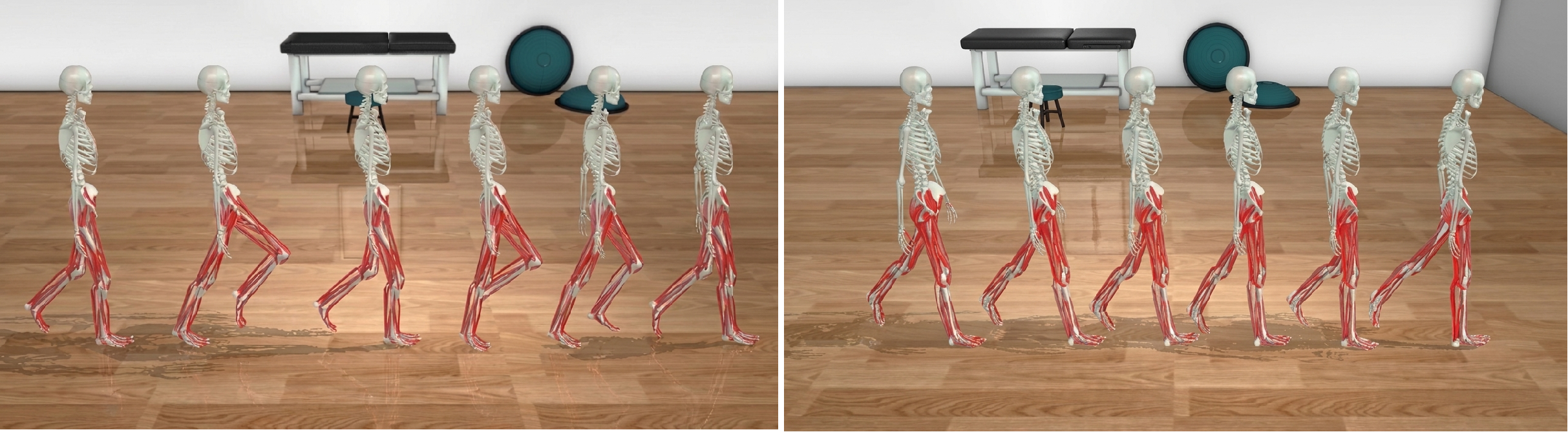}
    \caption{Qualitative comparison of evolved gaits: (Left) The \textit{full SDE} framework emerges a stable, alternating bipedal gait with natural leg-swing phases. (Right) The \textit{SDE-$\sigma$} (Strength-only) variant falls into a sub-optimal ``scissor-gait'', where the limbs remain in a crossed configuration and the agent moves via hopping.}
    \label{fig:gait_comparison}
\end{figure}

Video-based evaluation (Fig.~\ref{fig:gait_comparison}) indicates that the \textit{SDE-$\sigma$} model exhibits an unnatural ``scissor-gait'', where after the initial step, the musculoskeletal limbs remain in a crossed configuration, propelling the torso forward through synchronous hopping rather than coordinated, alternating strides. In contrast, the full SDE framework converges on a stable, alternating gait. This suggests that the three parameters do not contribute to locomotion in isolation but form a functional synergy. Specifically, while strength provides the necessary power, velocity ($\nu$) and stiffness ($\kappa$) are essential for regulating the temporal coordination and passive dynamics required for natural, human-like walking.

\begin{figure}[t]
    \centering
    \subfloat[$\sigma$ Evolution.]{
        \includegraphics[width=0.45\linewidth]{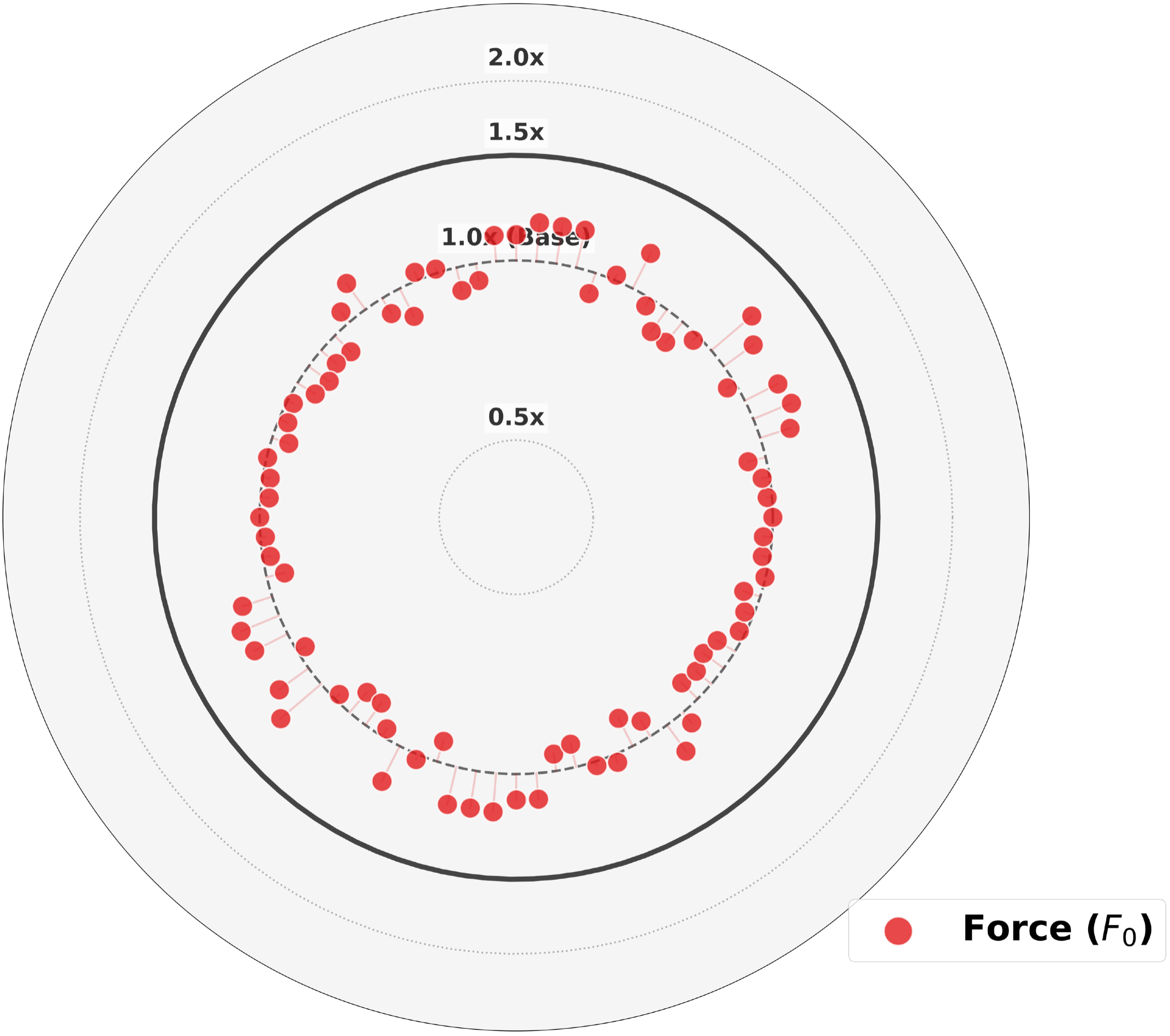}
        \label{fig:radar_strength}
    }
    \hfill
    \subfloat[$\nu$ Evolution.]{
        \includegraphics[width=0.45\linewidth]{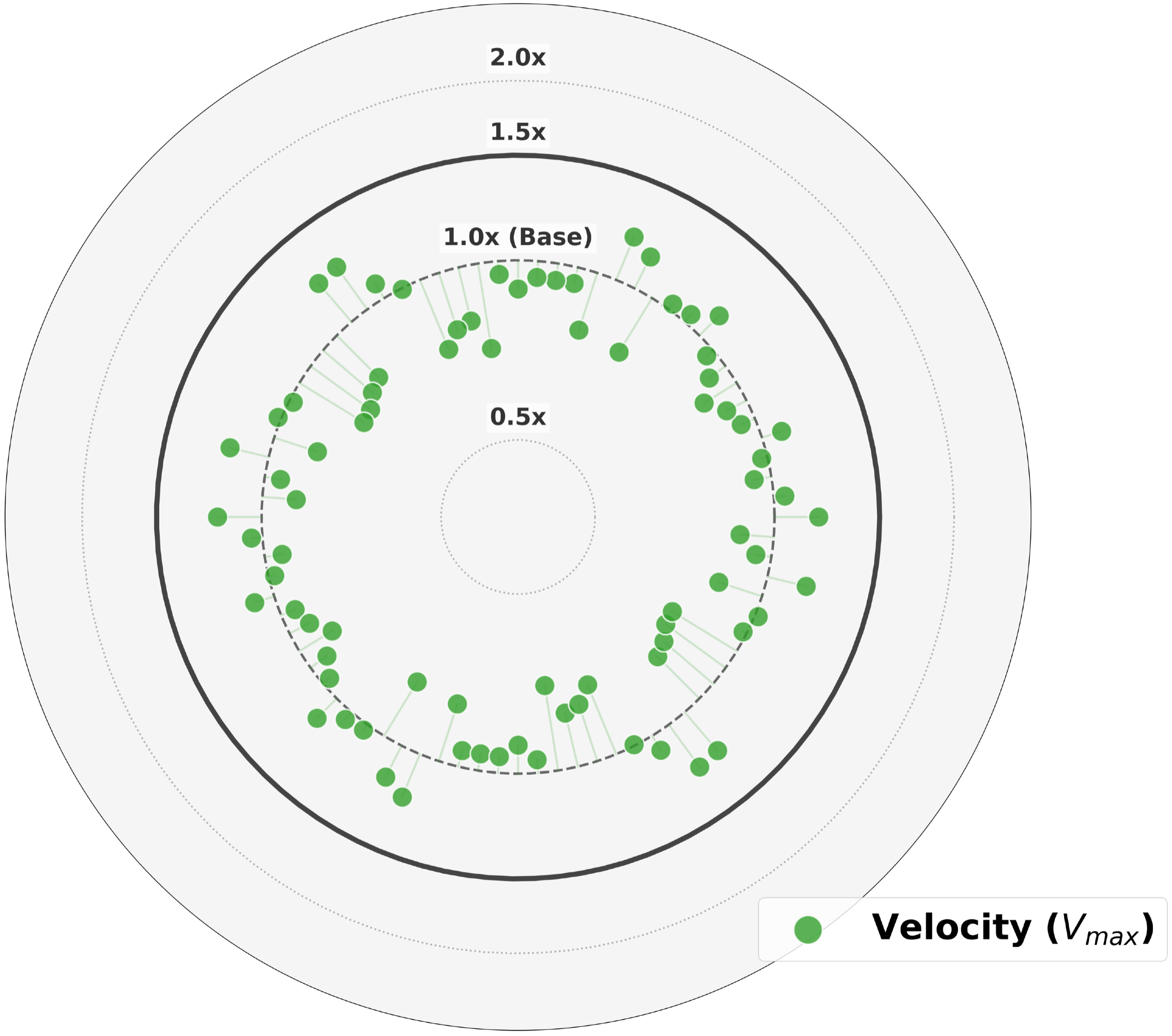}
        \label{fig:radar_velocity}
    }
    \\
    \subfloat[$\kappa$ Evolution.]{
        \includegraphics[width=0.45\linewidth]{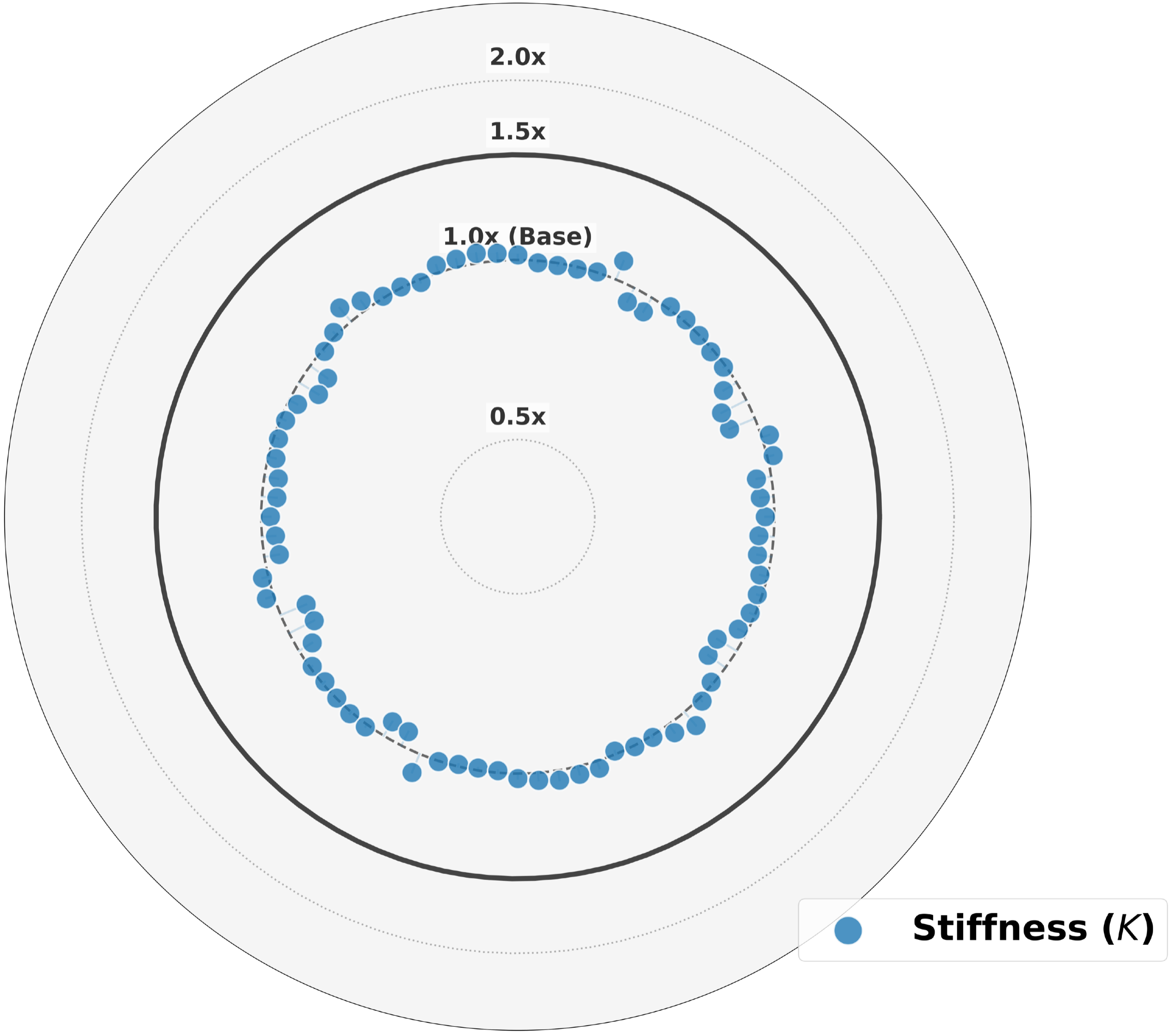}
        \label{fig:radar_stiffness}
    }
    \hfill
    \subfloat[SDE (Full Triad).]{
        \includegraphics[width=0.45\linewidth]{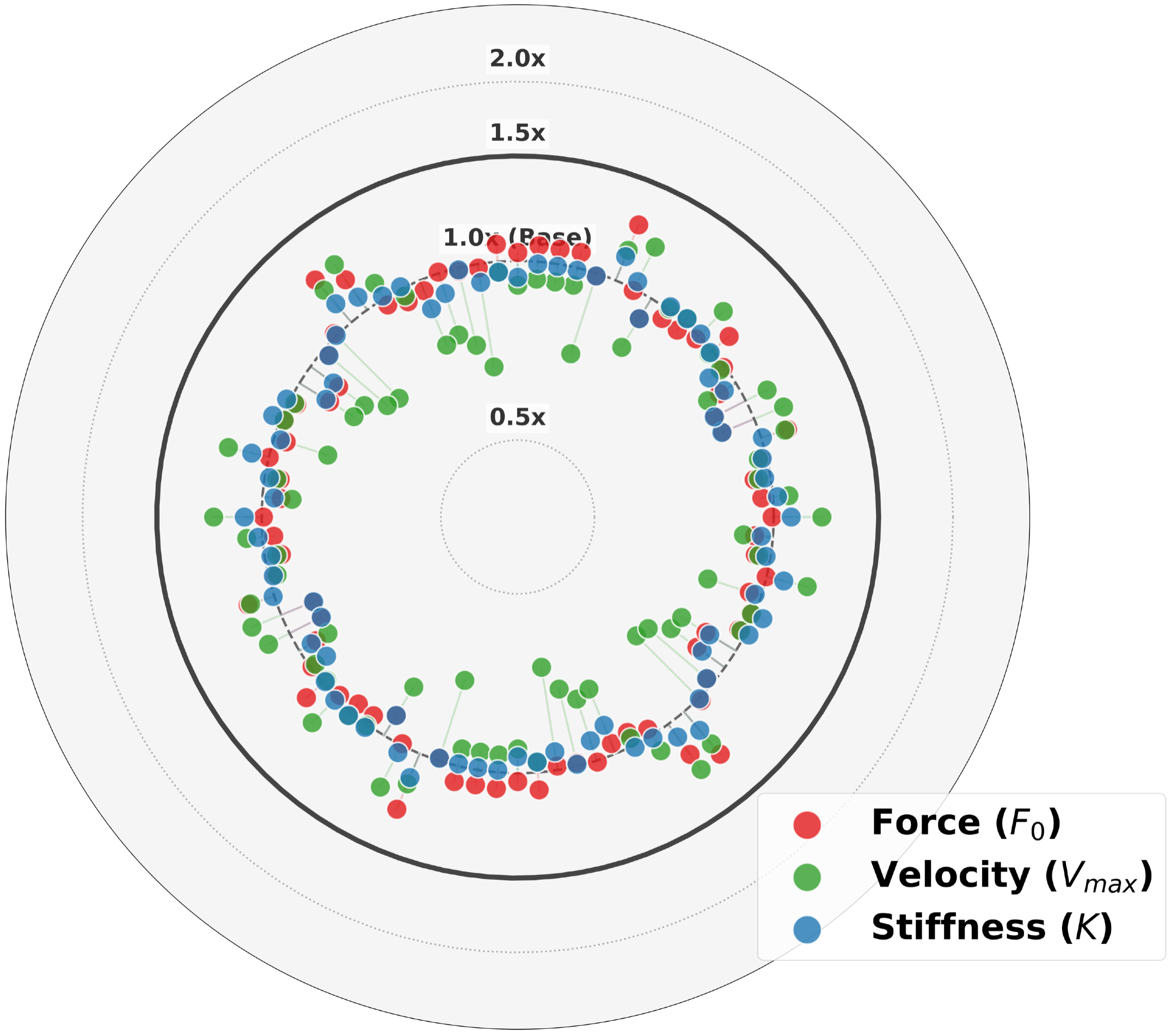}
        \label{fig:radar_pca}
    }
    \caption{Radar charts showing the distributions of evolved muscle parameters across different muscle groups. (a-c) depict the evolution of individual parameters: strength ($\sigma$), velocity ($\nu$), and stiffness ($\kappa$), which exhibit fragmented and inconsistent adaptation across muscles. (d) shows the SDE-evolved full triad, highlighting a coordinated synergy where muscle parameters co-adapt in a physiologically plausible manner, enabling efficient and stable locomotion.}
    \label{fig:radar_comparison}
\end{figure}

Figure~\ref{fig:radar_comparison} visualizes the distributions of evolved muscle parameters across different muscle groups. 
Subplots (a-c) show the single-parameter evolution of strength ($\sigma$), velocity ($\nu$), and stiffness ($\kappa$) respectively. 
We observe that optimizing parameters independently often leads to fragmented adaptation, with some muscle groups overcompensating while others remain under-optimized. 
In contrast, subplot (d) illustrates the SDE approach co-evolving the full triad ($\sigma, \nu, \kappa$). Here, muscle parameters exhibit a coherent, balanced pattern across groups, indicating physiologically plausible synergies that support more robust locomotion. 
These observations suggest that co-optimizing the triad helps avoid conflicting adaptations that may arise in single-parameter optimization.

\subsubsection{Impact of Bilateral Symmetry Prior}
\begin{figure}[t]
    \centering
    \subfloat[Walk Symmetry Ablation.]{
        \includegraphics[width=0.47\linewidth]{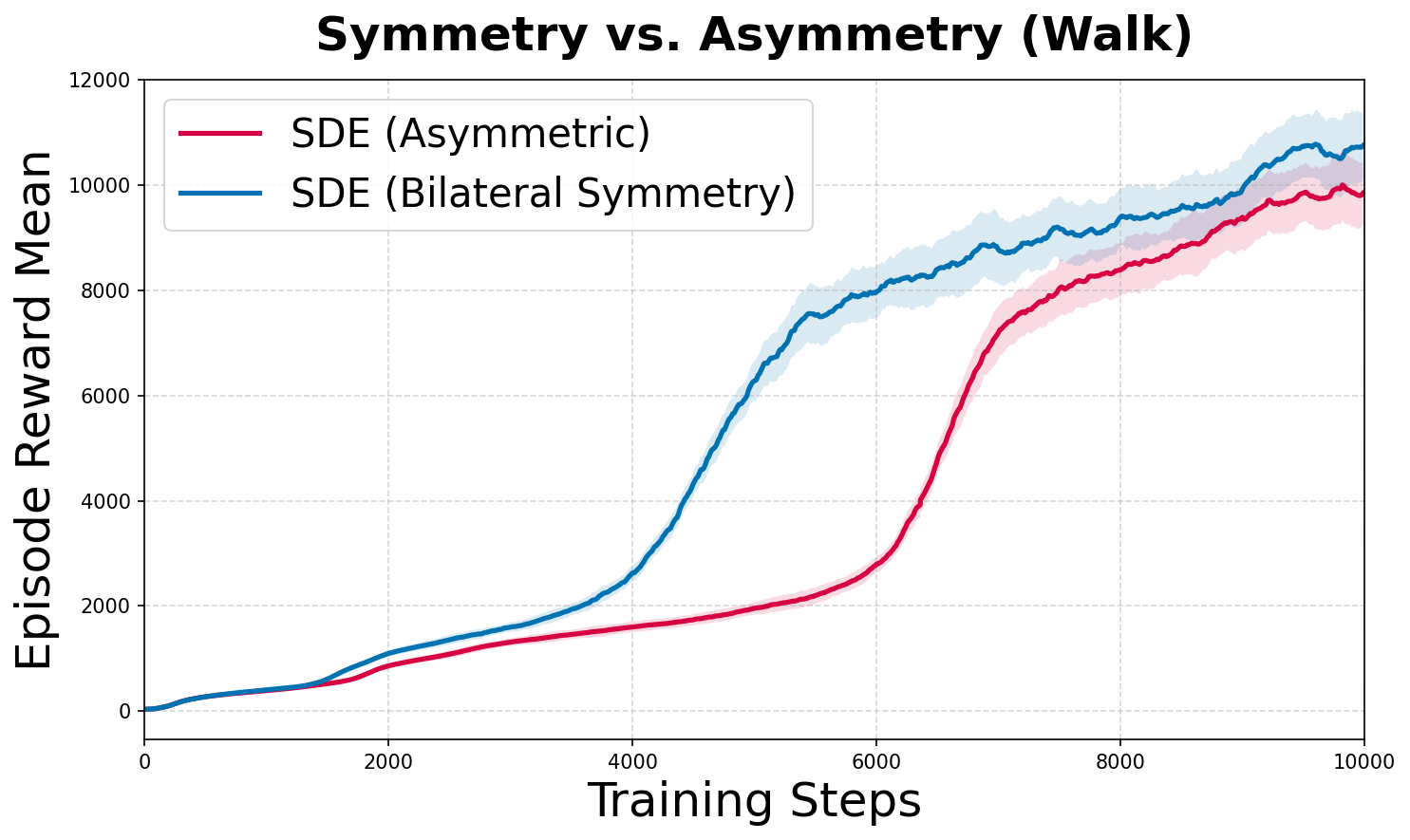}
        \label{fig:sym_walk}
    }
    \hfill
    \subfloat[Rough Symmetry Ablation.]{
        \includegraphics[width=0.47\linewidth]{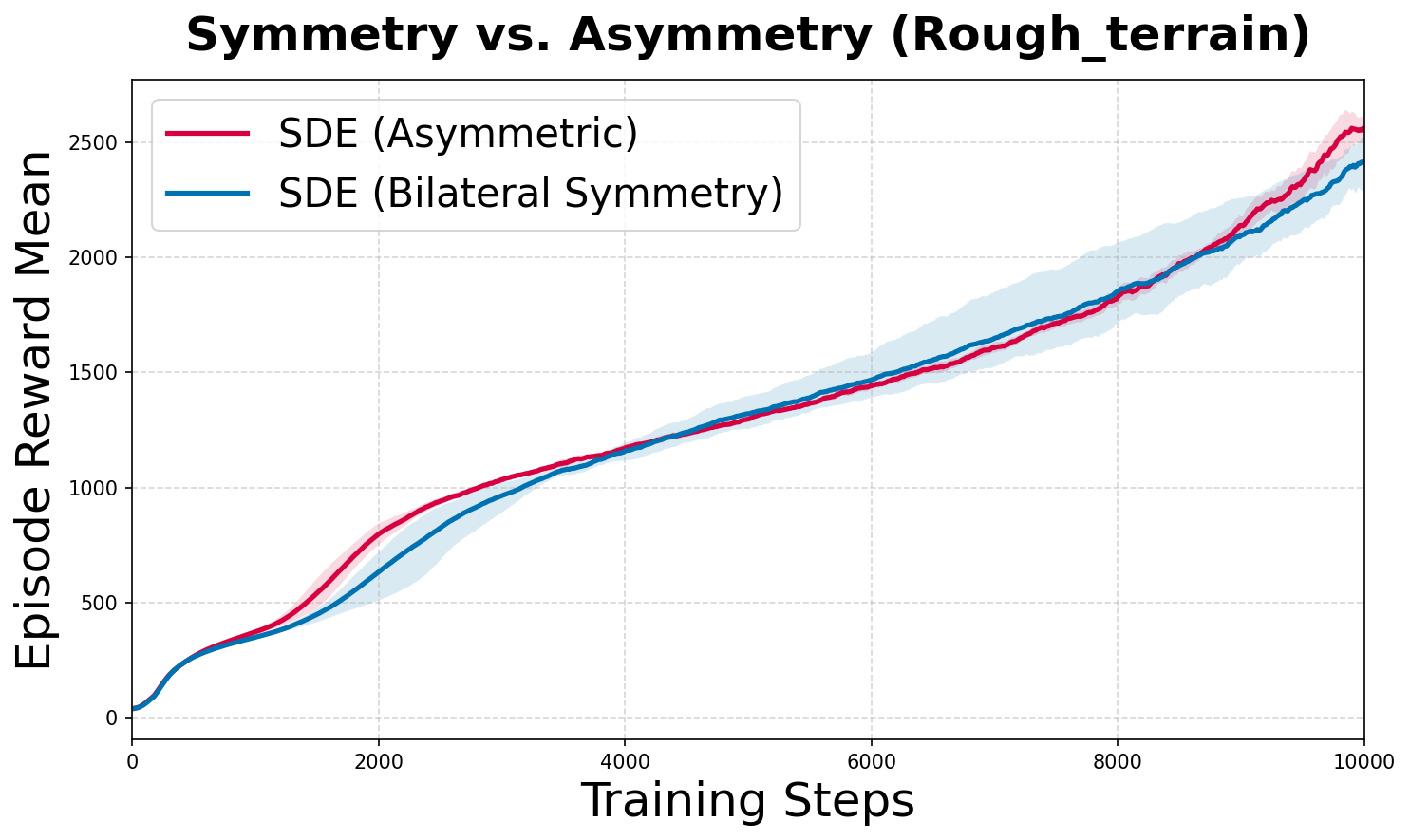}
        \label{fig:sym_rough}
    }
    \caption{Ablation study on bilateral symmetry. While \textit{SDE-Asym} has higher degrees of freedom, the symmetry prior in SDE ensures more stable and physically plausible gait patterns.}
    \label{fig:symmetry_ablation}
\end{figure}

To address \textbf{RQ3}, we compare the standard SDE (Symmetric) with an asymmetric variant (\textit{SDE-Asym}) in \textit{Walk} and \textit{Rough} tasks (Fig.~\ref{fig:symmetry_ablation}). 

\begin{itemize}
    \item \textbf{Regular Environments (Walk):} Enforcing bilateral symmetry provides a clear advantage. The symmetric variant converges faster and achieves a better final morphology. These results support the view that, for periodic and structured tasks, the symmetry prior serves as an effective inductive bias that aligns evolutionary exploration with balanced gait dynamics.
    \item \textbf{Irregular Environments (Rough):} In stochastic environments characterized by unpredictable rough terrain, the advantage of the bilateral symmetry prior is less pronounced. While the SDE with bilateral symmetry still maintains stability, the performance gap between symmetric and asymmetric configurations narrows. This suggests that in non-periodic, high-impact interactions, the ability to perform asymmetric morphological compensations may be more tolerable, or even necessary, to absorb stochastic perturbations.
\end{itemize}

\subsubsection{Impact of Latent Dimension $k$}
To address \textbf{RQ4}, we investigate the sensitivity of SDE to the latent dimension $k \in \{3, 5, 7, 9\}$, which corresponds to the number of principal components (PCs) retained in the spectral manifold. 
Before analyzing performance, we first construct the spectral manifold for each $k$. Specifically, we apply random perturbations to joint angles of the musculoskeletal robot and record the resulting muscle length trajectories across all muscles. 
This data captures the intrinsic dynamical correlations between muscles under arbitrary excitation. 
We then perform Principal Component Analysis (PCA) on the standardized muscle length history and retain the top $k$ eigenvectors, forming a low-dimensional spectral manifold that represents coordinated muscle co-variation.

\begin{figure}[t]
    \centering
    \includegraphics[width=0.95\linewidth]{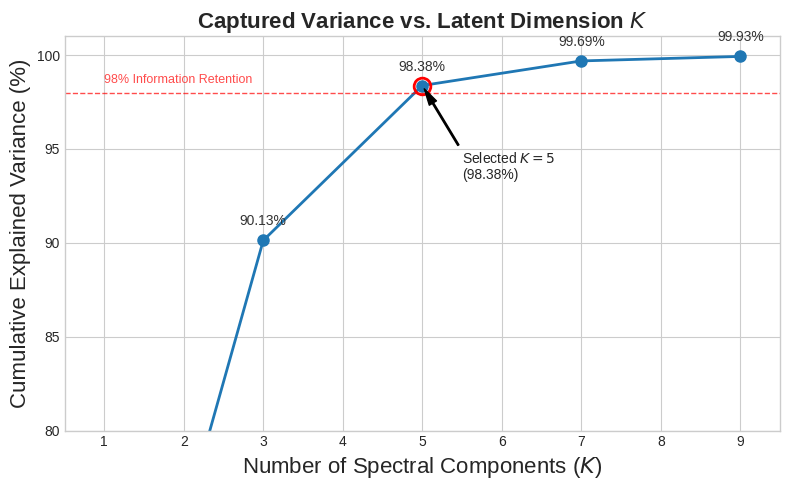}
    \caption{Cumulative explained variance of musculoskeletal dynamical response across different $k$. A distinct ``elbow'' is observed at $k=5$, capturing over $98\%$ of the morphological variance.}
    \label{fig:variance_scree}
\end{figure}

As illustrated in Fig.~\ref{fig:variance_scree}, the cumulative explained variance of the spectral manifold rises sharply, with $k=3$ already capturing $90.13\%$ of the original parameter space dynamics. A significant ``elbow'' effect occurs at $k=5$, where the captured variance reaches $98.38\%$. 

\begin{figure}[t]
    \vspace{10pt}
    \centering
    \includegraphics[width=0.95\linewidth] {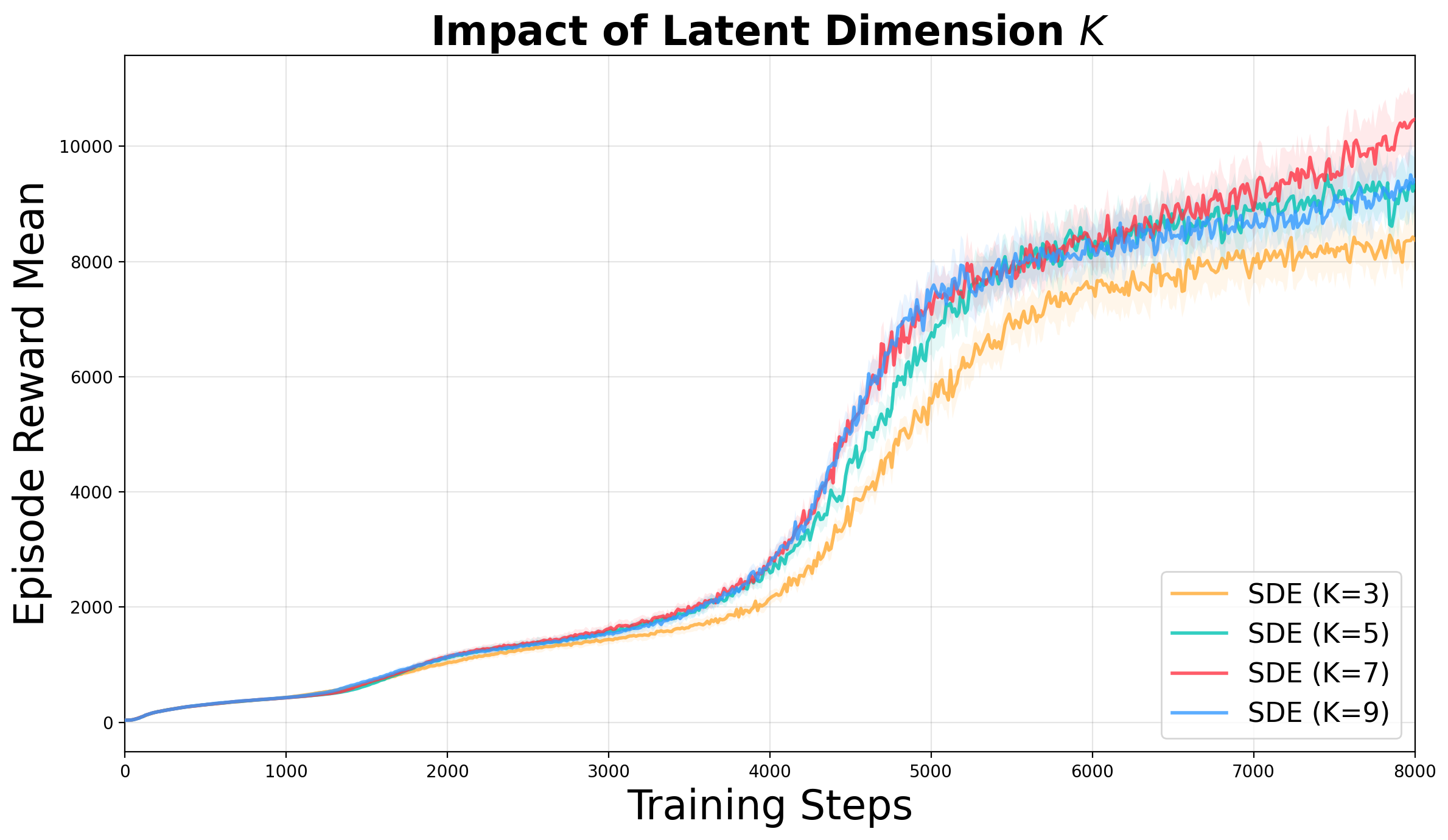}
    \caption{Learning curves on the \textit{Rough} task for different latent dimensions $k$. Although $k=7$ and $k=9$ offer higher theoretical expressivity, $k=5$ achieves the most efficient balance between convergence speed and final reward.}
    \label{fig:k_ablation}
\end{figure}

The impact of this dimensionality on actual task performance is shown in Fig.~\ref{fig:k_ablation}. We observe that while $k=7$ and $k=9$ eventually achieve slightly higher peak rewards, they exhibit slower initial convergence and higher variance across training seeds. Conversely, $k=3$ suffers from under-fitting. \textbf{Ultimately, $k=5$ is selected as the default configuration for SDE.}

\section{Conclusion}
In this paper, we introduced \textit{Spectral Design Evolution} (SDE), a co-optimization framework that accelerates musculoskeletal robot synthesis by evolving a triad of physiological muscle parameters: strength, velocity, and stiffness. To overcome the curse of dimensionality in high-dimensional muscle spaces, SDE leverages bilateral symmetry priors and PCA-based spectral manifolds to project complex morphological traits into a compact latent space. 

Experimental evaluations across diverse and challenging terrains demonstrate that SDE significantly outperforms fixed-morphology and standard co-optimization baselines in both convergence efficiency and locomotion stability. Ablation studies further validate that the synergistic evolution of the muscle triad, constrained by symmetry priors, is essential for achieving superior robustness. Future research will explore multi-task scenarios, enabling universal design policies to dynamically adapt morphology for real-time transitions between heterogeneous environments.

\bibliographystyle{IEEEtran}
\bibliography{ref}

@STRING{IEEE_J_RA_L         = "{IEEE} Robot. Autom. Lett."}

@inproceedings{vittori2022myosuite,
 author    = {Vittori, A. and Caggiano, V. and others},
 title     = {{{MyoSuite}}: A contact-rich simulation suite for musculoskeletal motor control},
 booktitle = {Proc. 5th Conf. Robot Learning (CoRL)},
 year      = {2022}
}

@article{schulman2017proximalpolicyoptimizationalgorithms,
 title     = {Proximal Policy Optimization Algorithms}, 
 author    = {Schulman, John and Wolski, Filip and Dhariwal, Prafulla and Radford, Alec and Klimov, Oleg},
 journal   = {arXiv preprint arXiv:1707.06347},
 year      = {2017}
}

@article{liu2025embodied,
 title     = {Embodied intelligence: A synergy of morphology, action, perception and learning},
 author    = {Liu, Huaping and Guo, Di and Cangelosi, Angelo},
 journal   = {ACM Comput. Surv.},
 volume    = {57},
 number    = {7},
 pages     = {1--36},
 year      = {2025}
}

@inproceedings{he2024dynsyn,
 author    = {He, Kaibo and Zuo, Chenhui and Sui, Yanan},
 title     = {{{DynSyn}}: Dynamical Synergistic Representation for Efficient Learning and Control},
 booktitle = {Proc. 41st Int. Conf. Mach. Learn. (ICML)},
 year      = {2024}
}

@article{schumacher2022dep,
 title     = {Dep-rl: Embodied exploration for reinforcement learning in overactuated and musculoskeletal systems},
 author    = {Schumacher, Pierre and H{\"a}ufle, Daniel and B{\"u}chler, Dieter and Schmitt, Syn and Martius, Georg},
 journal   = {arXiv preprint arXiv:2206.00484},
 year      = {2022}
}

@inproceedings{zhao2023bayesian,
 author    = {Zhao, J. and Yang, Y. and Liu, H.},
 title     = {Bayesian Morphology Optimization for Musculoskeletal Systems},
 booktitle = {Proc. IEEE/RSJ Int. Conf. Intell. Robots Syst. (IROS)},
 year      = {2023}
}

@inproceedings{yuan2022t2a,
 author    = {Yuan, Y. and Song, Y. and Kitani, K.},
 title     = {{{Transform2Act}}: Learning a Transform-and-Control Policy for Efficient Agent Design},
 booktitle = {Proc. 10th Int. Conf. Learn. Represent. (ICLR)},
 year      = {2022}
}

@article{zajac1989muscle,
 author    = {F. E. Zajac},
 title     = {Muscle and tendon: properties, models, scaling, and application to biomechanics},
 journal   = {Crit. Rev. Biomed. Eng.},
 year      = {1989},
 volume    = {17},
 number    = {4},
 pages     = {359--411}
}

@article{santello1998postural,
 author    = {M. Santello and M. Flanders and J. F. Soechting},
 title     = {Postural hand synergies for tool use},
 journal   = {J. Neurosci.},
 year      = {1998},
 volume    = {18},
 pages     = {10105--10115}
}

@inproceedings{caggiano2022myosuite,
 author    = {Caggiano, V. and others},
 title     = {{{MyoSuite}}: A Platform for Biomechanics with Deep Reinforcement Learning},
 booktitle = {Proc. Int. Conf. Mach. Learn. (ICML)},
 year      = {2022}
}

@inproceedings{sims1994evolving,
 author    = {K. Sims},
 title     = {Evolving {3D} Morphology and Behavior by Competition},
 booktitle = {Artificial Life IV},
 pages     = {28--39},
 year      = {1994}
}

@book{pfeifer2006body,
 author    = {R. Pfeifer and J. Bongard},
 title     = {How the Body Shapes the Way We Think},
 publisher = {MIT Press},
 year      = {2006}
}

@article{scott2004optimal,
 author    = {S. H. Scott},
 title     = {Optimal feedback control and the neural control of movement},
 journal   = {Nature Rev. Neurosci.},
 year      = {2004},
 volume    = {5},
 pages     = {532--546}
}

@article{seth2018opensim,
 author    = {Seth, A. and others},
 title     = {{{OpenSim}}: Simulating musculoskeletal dynamics and neuromuscular control},
 journal   = {PLoS Comput. Biol.},
 year      = {2018},
 volume    = {14}
}

@article{peng2018deepmimic,
 author    = {X. B. Peng and others},
 title     = {{{DeepMimic}}: Example-guided deep reinforcement learning of physics-based character skills},
 journal   = {ACM Trans. Graph.},
 year      = {2018}
}

@article{d2003combinations,
 author    = {d'Avella, G. and others},
 title     = {Combinations of muscle synergies in the construction of a natural motor repertoire},
 journal   = {Nature Neurosci.},
 year      = {2003},
 volume    = {6}
}

@article{chiappa2024acquiring,
 author    = {A. S. Chiappa and others},
 title     = {Acquiring musculoskeletal skills with a hierarchical reinforcement learning framework},
 journal   = {Neuron},
 year      = {2024}
}

@inproceedings{luck2020data,
 author    = {K. S. Luck and J. Amor and H. B. Amor},
 title     = {Data-efficient co-adaptation of morphology and control with adaptive inference},
 booktitle = {Proc. IEEE Int. Conf. Robot. Autom. (ICRA)},
 year      = {2020},
 pages     = {2222--2228}
}

@article{simpson2021data,
  title     = {Data-Driven Spectral Submanifold Reduction for Nonlinear Optimal Control of High-Dimensional Robots},
  author    = {Simpson, Thomas and Cenedese, Mattia and Axas, Ioannis and Haller, George},
  journal   = IEEE_J_RA_L,
  volume    = {6},
  number    = {3},
  pages     = {5304--5311},
  year      = {2021}
}

@article{diffmuscle2026,
  author    = {Zhao, Wentao and Guo, Jun and Huang, Kangyao and Liu, Xin and Liu, Huaping},
  title     = {Diff-Muscle: Efficient Learning for Musculoskeletal Robotic Table Tennis},
  journal   = {arXiv preprint arXiv:2603.08617},
  year      = {2026}
}

@inproceedings{dong2023sard,
  title     = {Symmetry-Aware Robot Design with Structured Subgroups},
  author    = {Dong, Heng and Zhang, Junyu and Wang, Tonghan and Zhang, Chongjie},
  booktitle = {Proc. 40th Int. Conf. Mach. Learn. (ICML)},
  year      = {2023}
}

@article{huang2024competevo,
  title     = {CompetEvo: Towards Morphological Evolution from Competition},
  author    = {Huang, Kangyao and Guo, Di and Zhang, Xinyu and Ji, Xiangyang and Liu, Huaping},
  journal   = {arXiv preprint arXiv:2405.18300},
  year      = {2024}
}

@inproceedings{lu2025bodygen,
  title     = {{{BodyGen}}: Advancing Towards Efficient Embodiment Co-Design},
  author    = {Lu, Haofei and Wu, Zhe and Xing, Junliang and Li, Jianshu and Li, Ruoyu and Li, Zhe and Shi, Yuanchun},
  booktitle = {Proc. 13th Int. Conf. Learn. Represent. (ICLR)},
  year      = {2025}
}

@article{millard2013flexing,
  author    = {Millard, M. and Uchida, T. and Seth, A. and Delp, S. L.},
  title     = {Flexing Computational Muscle: Modeling and Simulation of Musculotendon Dynamics},
  journal   = {J. Biomech. Eng.},
  volume    = {135},
  number    = {2},
  pages     = {021005},
  year      = {2013}
}

\end{document}